\documentclass[letterpaper, 10 pt, conference]{ieeeconf}  

\IEEEoverridecommandlockouts             

\usepackage{graphics}
\usepackage{graphicx}
\usepackage{amsmath}
\usepackage{amssymb}
\usepackage{bm}
\usepackage{multirow}
\usepackage{color}
\usepackage{booktabs}
\usepackage{pifont}
\newcommand{\cmark}{\ding{51}}
\newcommand{\xmark}{\ding{55}}

\definecolor{Red}{cmyk}{0,1,1,0}
\definecolor{Green}{cmyk}{1,0,1,0}
\definecolor{Cyan}{cmyk}{1,0,0,0}
\definecolor{Purple}{cmyk}{0.45,0.86,0,0}
\definecolor{Rosolic}{cmyk}{0.00,1.00,0.50,0}
\definecolor{Blue}{cmyk}{1.00,1.00,0.00,0}
\definecolor{Orange}{cmyk}{0,0.52,0.80,0}
\definecolor{Black}{cmyk}{1,0,0,1}

\title{\LARGE \bf
SemGS: Feed-Forward Semantic 3D Gaussian Splatting from Sparse Views for Generalizable Scene Understanding
}

\author{Sheng Ye$^{1}$, Zhen-Hui Dong$^{1}$, Ruoyu Fan$^{1}$, Tian Lv$^{1}$, and Yong-Jin Liu$^{1,*}$,~\IEEEmembership{Senior Member,~IEEE}%
\thanks{$^{1}$S. Ye, Z.-H. Dong, R. Fan, T. Lv, and Y.-J. Liu are with the Department of Computer Science, Tsinghua University, Beijing, China {\tt\small \{yec22, dzh23, fry21, lt22\}@mails.tsinghua.edu.cn}, \tt\small liuyongjin@tsinghua.edu.cn}%
\thanks{*Corresponding Author}%
}

\begin{document}

\maketitle
\thispagestyle{empty}
\pagestyle{empty}


\begin{abstract}
Semantic understanding of 3D scenes is essential for robots to operate effectively and safely in complex environments. Existing methods for semantic scene reconstruction and semantic-aware novel view synthesis often rely on dense multi-view inputs and require scene-specific optimization, limiting their practicality and scalability in real-world applications. To address these challenges, we propose \textbf{SemGS}, a feed-forward framework for reconstructing generalizable semantic fields from sparse image inputs. \textbf{SemGS} uses a dual-branch architecture to extract color and semantic features, where the two branches share shallow CNN layers, allowing semantic reasoning to leverage textural and structural cues in color appearance. We also incorporate a camera-aware attention mechanism into the feature extractor to explicitly model geometric relationships between camera viewpoints. The extracted features are decoded into dual-Gaussians that share geometric consistency while preserving branch-specific attributes, and further rasterized to synthesize semantic maps under novel viewpoints. Additionally, we introduce a regional smoothness loss to enhance semantic coherence. Experiments show that \textbf{SemGS} achieves state-of-the-art performance on benchmark datasets, while providing rapid inference and strong generalization capabilities across diverse synthetic and real-world scenarios.
\end{abstract}

\section{INTRODUCTION}
Semantic understanding of 3D scenes is a fundamental challenge in computer vision and robotics. For intelligent robots to operate safely and efficiently in unknown environments, they must go beyond low-level appearance perception and gain a high-level semantic understanding of their surroundings. Such semantic awareness is crucial for tasks like navigation~\cite{navigation, context-navigation}, obstacle avoidance~\cite{obstacle}, and decision-making~\cite{decision}. While recent advances in 3D scene representation — such as Neural Radiance Fields (NeRF)~\cite{nerf} and 3D Gaussian Splatting (3DGS)~\cite{3dgs} — have achieved remarkable rendering fidelity, they only provide implicit geometry and appearance details, without semantic reasoning.
Thus, there is a growing need to integrate semantic information into these 3D representations. Despite its importance, semantic scene understanding and semantic-aware novel view synthesis under sparse inputs remain under-explored.

Some pioneering works~\cite{SemanticNeRF, LERF, Langsplat} have extended NeRF and 3DGS to incorporate semantics, either through auxiliary semantic branches or embedding semantic features~\cite{clip, LSeg}.
However, these methods typically rely on dense multi-view images, which are costly to acquire. Furthermore, they are generally optimized in a scene-specific manner. For each new scenario, these methods have to retrain a new model, which severely limits their scalability and real-world applicability.

\begin{figure}[t]
  \centering
  \setlength{\abovecaptionskip}{0pt}  
  \setlength{\belowcaptionskip}{0pt}
  \includegraphics[width=\linewidth]{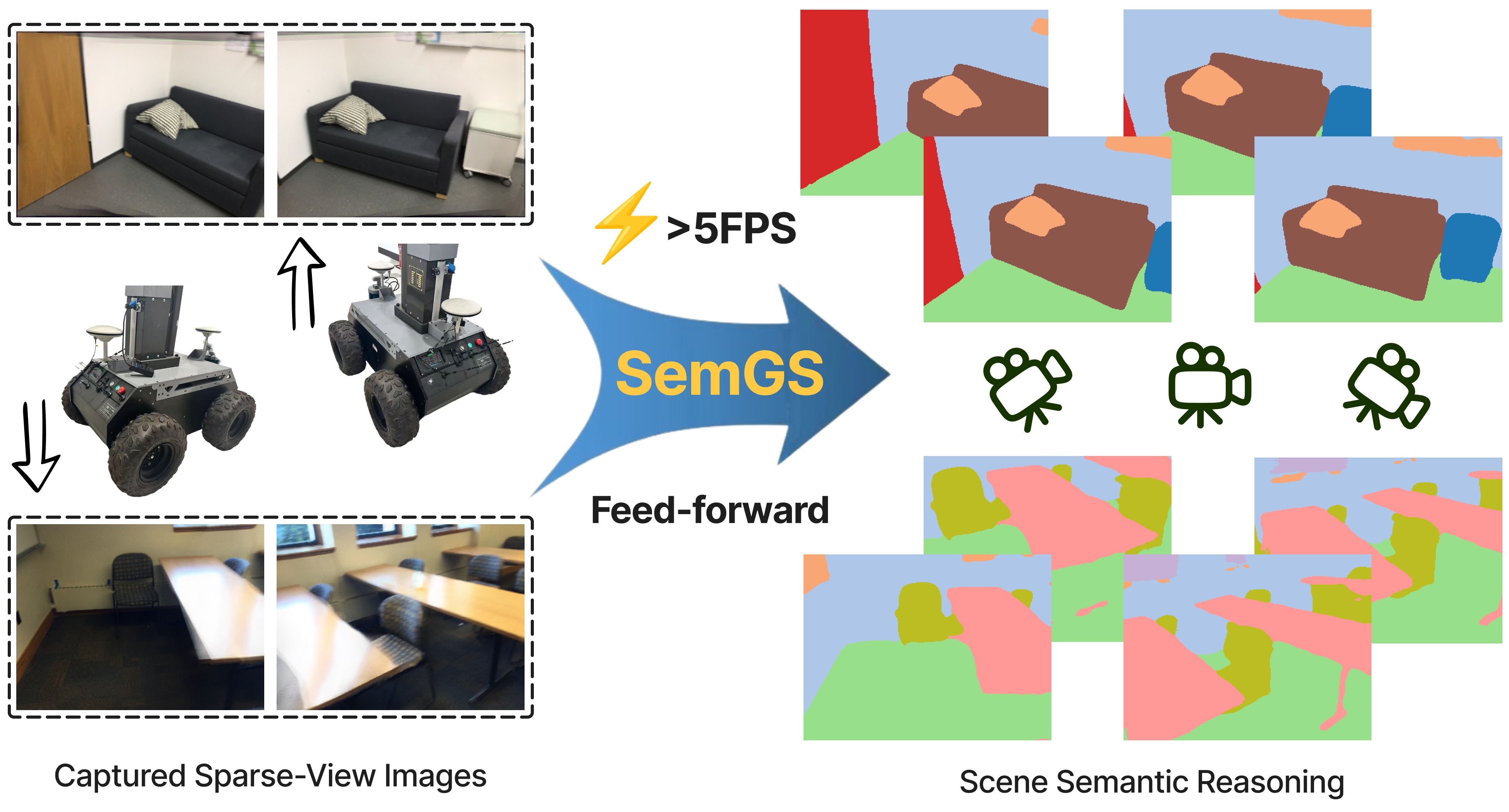}
  \caption{We propose \textbf{SemGS}, a novel framework for generalizable semantic 3DGS. Given sparse-view images of an unseen scene, \textbf{SemGS} can rapidly infer semantic maps under novel viewpoints in a single feed-forward pass.}
  \label{fig:teaser}
\end{figure}

In this work, we aim to learn a generalizable semantic representation that can be trained across multiple scenes and infer semantic maps under novel viewpoints from only sparse input images. As shown in Fig.~\ref{fig:teaser}, our method enables fast semantic inference in a single feed-forward pass.
Achieving this capability requires the model to possess strong geometric reasoning and generalization abilities, rather than merely overfitting to a specific scene.
Our work is motivated by recent feed-forward 3DGS methods such as MVSplat~\cite{mvsplat}, which leverages neural networks to predict Gaussian attributes from sparse views in a feed-forward pass. While these feed-forward methods focus solely on color rendering, we observe that through cost-volume based depth reasoning, such models inherently capture rich geometric priors that are also highly informative for semantic inference. Since visual appearance and semantics are often closely related, extending the feed-forward 3DGS from color to semantic can be both intuitive and beneficial.

To achieve this goal, we propose \textbf{SemGS}, a novel feed-forward framework for generalizable semantic 3DGS.
Our method employs dual branches for color and semantic feature extraction, each consisting of a CNN backbone and a Swin Transformer~\cite{swin}, with cross-attention mechanisms fusing information across multiple input views.
By sharing low-level CNN feature layers between the two branches, \textbf{SemGS} enables semantic reasoning to leverage textural and structural cues embedded in appearance representations.
Both color and semantic features are decoded into dual-Gaussians (\textit{color Gaussians} and \textit{semantic Gaussians}), whose geometric position and opacity attributes are shared and derived from the cost-volume based depth.
This design allows semantic Gaussians to inherit strong 3D geometric priors from the color reconstruction branch.
These Gaussians are then splatted~\cite{3dgs} to render novel views and semantic maps.

Inspired by PRoPE~\cite{prope}, we integrate camera intrinsic and extrinsic parameters into the Swin Transformer’s attention blocks via relative positional encoding, which enhances the 3D geometric awareness. We further introduce a regional smoothness loss to promote local semantic label consistency across neighboring regions.
Experiments show that \textbf{SemGS} achieves state-of-the-art performance on the ScanNet~\cite{scannet} and ScanNet++~\cite{scannet++} datasets, and generalizes robustly across synthetic (Replica~\cite{replica}) and real-world scenarios.

\begin{figure*}
    \centering
    \includegraphics[width=\linewidth]{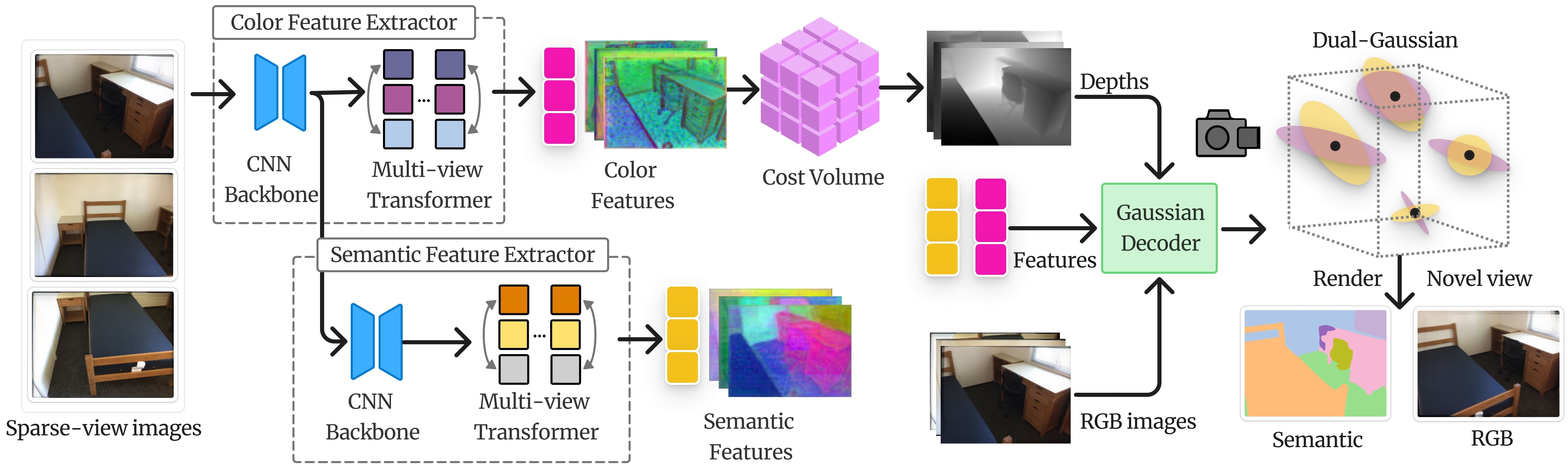}
    \caption{Overall pipeline of our proposed \textbf{SemGS}. Given sparse-view RGB images as inputs, \textbf{SemGS} leverages a dual-branch architecture (Sec.~\ref{sec:dual-branch-feature}) to extract both color and semantic features. These features are used to regress multi-view depth maps (Sec.~\ref{sec:depth}) and are subsequently decoded into per-pixel dual-Gaussians (Sec.~\ref{sec:gaussian-prediction}). The resulting dual-Gaussians share geometric attributes while maintaining branch-specific attributes, enabling efficient rasterization for synthesizing both novel RGB views and semantic maps.}
    \label{fig:pipeline}
\end{figure*}

In summary, our contributions are:
\begin{itemize}
\item We propose \textbf{SemGS}, a novel feed-forward framework for joint radiance and semantic field reconstruction from sparse input images, enabling rapid semantic inference without the need for per-scene optimization.
\item We incorporate camera geometry into the Swin Transformer via relative positional encoding to facilitate 3D perception, and introduce a regional smoothness loss to enforce semantic coherence.
\item Experiments show that \textbf{SemGS} outperforms existing baselines on benchmark datasets, while offering faster inference speed and stronger generalization ability.
\end{itemize}

\section{RELATED WORKS}
\subsection{Generalizable Novel View Synthesis}
The novel view synthesis task aims to generate photo-realistic images from previously unseen camera viewpoints.
NeRF~\cite{nerf} represents a significant breakthrough, modeling 3D scenes as continuous implicit functions encoded by MLPs and synthesizing novel views through volume rendering with high fidelity. Following this, 3D Gaussian Splatting (3DGS)~\cite{3dgs} further improves rendering efficiency using a differentiable point-based splatting technique.
However, both NeRF and 3DGS generally require per-scene optimization, which limits their scalability and generalizability.

To achieve cross-scene generalization, pixelNeRF~\cite{pixelnerf} conditions the radiance field on input image features, enabling synthesis from sparse input views. Subsequent approaches such as MVSNeRF~\cite{mvsnerf} and GeoNeRF~\cite{geonerf} integrate multi-view stereo and geometric priors to enhance rendering quality.
As 3DGS offers superior rendering quality and efficiency over NeRF, its generalizable extensions (a.k.a. \textit{feed-forward 3DGS}) have attracted growing interest. For instance, PixelSplat~\cite{pixelsplat} introduces a multi-view epipolar transformer to predict pixel-aligned Gaussian parameters in a feed-forward manner, while MVSplat~\cite{mvsplat} leverages cost volumes for improved geometry estimation.
Despite these advances, existing methods focus primarily on color rendering, leaving the problem of generalizable semantic reasoning largely unaddressed.

\subsection{Semantic Fields and Scene Understanding}
Semantic scene understanding is essential for many vision and robotic applications, including navigation~\cite{context-navigation} and human–robot interaction~\cite{interaction}, as it provides interpretable, high-level cues beyond raw appearance.
Conventional 3D semantic segmentation methods operate on point clouds~\cite{superpoint, graph} or meshes~\cite{virtual, vmnet}, but require expensive 3D annotations for training.
Recently, 3D semantic field methods have emerged that allow semantic map rendering from arbitrary viewpoints using only 2D supervision.
Specifically, Semantic-NeRF~\cite{SemanticNeRF} extends the radiance field with a semantic branch to jointly render color and semantic labels. Panoptic NeRF~\cite{panoptic} introduces a 3D-to-2D label transfer strategy to optimize both semantic and radiance fields. LERF~\cite{LERF} incorporates CLIP~\cite{clip} features into NeRF for open-vocabulary semantic segmentation.
In the context of 3DGS, several methods have been proposed: LangSplat~\cite{Langsplat} leverages SAM~\cite{sam} and CLIP embeddings to learn hierarchical semantic fields, Feature 3DGS~\cite{feature3dgs} enables semantic rendering via 3D feature field distillation, and Semantic Gaussians~\cite{semanticgaussians} introduce a 3D semantic network to associate semantic attributes with each Gaussian. Nevertheless, these methods are confined to scene-specific optimization and cannot generalize across scenes.

Research on generalizable semantic fields is still relatively limited. S-Ray~\cite{S-Ray} learns a generalizable semantic field by building a 3D context space with cross-reprojection attention. GSNeRF~\cite{GSNeRF} aggregates visual features and depth estimates to render novel-view semantic segmentations.
In contrast to these works, we propose a novel framework based on feed-forward 3DGS that efficiently constructs generalizable semantic fields from sparse-view inputs, achieving superior performance with faster inference speed.

\section{METHOD}
We propose \textbf{SemGS}, a feed-forward framework for reconstructing 3D scenes with semantic understanding from sparse input views.
Given \( N \) sparse-view RGB images \( \mathcal{I} = \{\mathbf{I}_i \in \mathbb{R}^{H \times W} \}_{i = 1}^N \), and their corresponding camera poses \( \mathcal{P} = \{\mathbf{K}_i,\mathbf{R}_i, \mathbf{t}_i\}_{i = 1}^N \), our model predicts a set of Gaussian primitives that jointly represent the geometry, radiance, and semantics of the 3D scene.

A key aspect of our method is the use of a \textbf{dual-Gaussian representation}, in which each pixel of the input images is associated with two complementary Gaussians: a \textit{color Gaussian} for radiance modeling and a \textit{semantic Gaussian} for semantic reasoning.
To exploit geometric priors and ensure consistency, both Gaussians share the same 3D position $\bm{\mu}_j$ and opacity $\alpha_j$, while maintaining their own branch-specific attributes: $(\mathbf{\Sigma}^c_j,\mathbf{c}_j)$ for color Gaussians and $(\mathbf{\Sigma}^s_j,\mathbf{s}_j)$ for semantic Gaussians, where $\mathbf{\Sigma}^c_j,\mathbf{\Sigma}^s_j$ denote the covariance matrices, $\mathbf{c}_j$ is the color encoded via spherical harmonics, and $\mathbf{s}_j$ is the semantic class label distribution.

As shown in Fig.~\ref{fig:pipeline}, our architecture comprises two parallel feature extraction branches (one for color and one for semantics), followed by a Gaussian decoder that outputs shared geometric attributes and branch-specific attributes.
Finally, the predicted Gaussians are rendered into novel views and semantic maps via a differentiable rasterizer~\cite{3dgs}.

\subsection{Multi-View Feature Extraction}
\label{sec:dual-branch-feature}
\subsubsection{Dual-Branch Feature Extractor}
We extract multi-view features through a dual-branch architecture tailored for capturing color and semantic information.
Both branches share the low-level CNN layers to capture fundamental texture and structure patterns, while maintaining branch-specific Swin Transformers~\cite{swin} for high-level feature learning.
Concretely, given input images $\{\mathbf{I}_i\}_{i=1}^N$, the low-level features are first extracted by a shared CNN backbone.
In the color branch, these features are directly processed by a color Transformer to obtain per-view color features $\{\mathbf{F}^c_i\}_{i=1}^N$.
In parallel, the semantic branch employs an additional CNN to refine the features into semantic cues, which are subsequently fed into a semantic Transformer to yield per-view semantic features $\{\mathbf{F}^s_i\}_{i=1}^N$.
This design enables both branches to exploit shared low-level textural and structural information while learning branch-specific high-level features.

\subsubsection{Camera Pose Injection}
\begin{figure}
    \centering
    \includegraphics[width=0.9\linewidth]{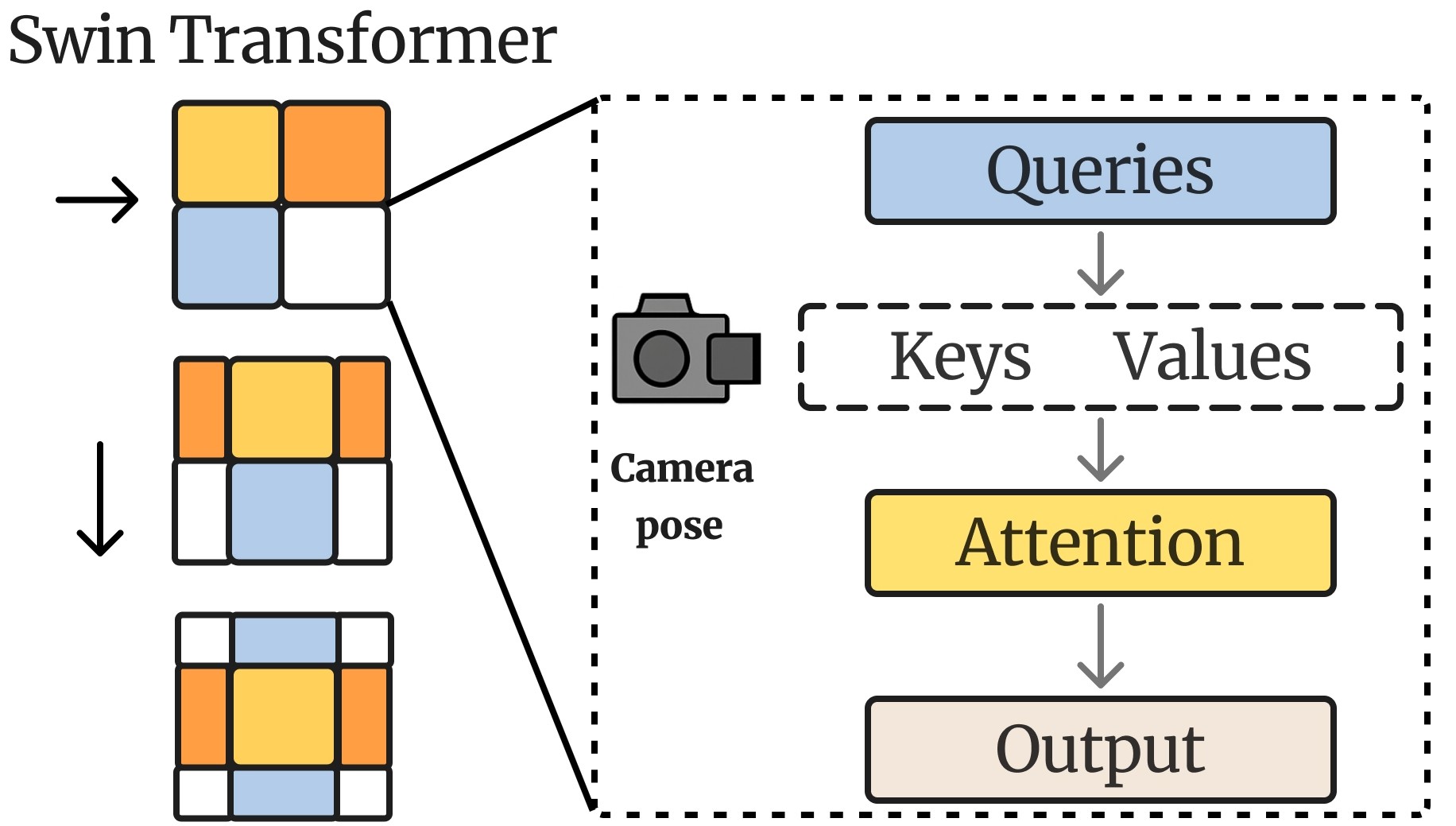}
    \caption{Swin Transformer with camera-aware attention. Different colors denote attention windows, and the window shifting mechanism enables model to capture cross-window connections. Camera poses are injected into the attention process to enhance the 3D reasoning ability.}
    \label{fig:swin_transformer}
\end{figure}

Our dual-branch feature extractor utilizes Swin Transformer to learn high-level features through its shifted window mechanism, which effectively captures both local and global context. To further enhance inter-view consistency, Swin Transformers are extended with cross-attention layers that propagate information across multiple input views, enabling powerful 3D geometric reasoning.
Accurate geometric and semantic perception relies on camera information, particularly under sparse-view inputs. However, the original Swin Transformer struggles to capture geometric relationships across camera viewpoints.
Therefore, we propose to adopt a camera-aware attention mechanism (shown in Fig.~\ref{fig:swin_transformer}) inspired by PRoPE~\cite{prope}.
Specifically, we inject the camera poses (\textit{i.e.,} projective transformations) into attention process by applying encoding to the queries, keys, and values of tokens (\textit{i.e.,} image patch embeddings) in Swin Transformer.
Formally, the relative camera projective transformation between view $i$ and $j$ is:
\begin{equation}
\tilde{\mathbf{P}}_{i\to j}=\tilde{\mathbf{P}}_{i}\tilde{\mathbf{P}}_{j}^{-1},\ 
\tilde{\mathbf{P}}_{i}=\begin{bmatrix}
\mathbf{K}_i\mathbf{R}_i & \mathbf{K}_i\mathbf{t}_i\\
\mathbf{0}^\top & 1
\end{bmatrix}, 
\end{equation}
where $\tilde{\mathbf{P}}$ denotes the world-to-camera projection matrix, $\mathbf{K}_i$, $\mathbf{R}_i$ and $\mathbf{t}_i$ represent the camera intrinsic, rotation, and translation of view $i$, respectively.
For a token $t$, to jointly encode the inter-view relations and intra-view token positional order, we construct a block-diagonal matrix $\mathbf{G}_{t} \in \mathbb{R}^{d \times d}$:
\begin{equation}
\mathbf{G}_{t}=
\begin{bmatrix}
\mathbf{G}^{\mathrm{proj}}_{t} & 0\\
0 & \mathbf{G}^{\mathrm{rope}}_{t}
\end{bmatrix}.
\end{equation}
In this block-diagonal matrix $\mathbf{G}_{t}$, $\mathbf{G}^{\mathrm{proj}}_{t}=\mathbf{I}_{d/8}\otimes\tilde{\mathbf{P}}_{i(t)} \in \mathbb{R}^{\frac{d}{2} \times \frac{d}{2}}$ encodes camera-level information, and
\begin{equation}
\mathbf{G}^{\mathrm{rope}}_{t}=
\begin{bmatrix}
\mathrm{RoPE}(x_t) & \mathbf{0}\\
\mathbf{0} & \mathrm{RoPE}(y_t)
\end{bmatrix} \in \mathbb{R}^{\frac{d}{2} \times \frac{d}{2}}
\end{equation}
encodes intra-view positional information via rotary embeddings~\cite{rope}. Here, $\mathrm{RoPE}(\cdot)$ represents $\frac{d}{4}\times \frac{d}{4}$ rotary position embedding for the $(x_t, y_t)$ coordinate of token $t$.
Following the GTA-style attention~\cite{GTA}, $\mathbf{G}_{t}$ is then applied to the query, key, and value of token $t$ as:
\begin{equation}
\mathbf{Q}'_{t}=(\mathbf{G}_{t})^\top\mathbf{Q}_{t},\ 
\mathbf{K}'_{t}=(\mathbf{G}_{t})^{-1}\mathbf{K}_{t},\ 
\mathbf{V}'_{t}=(\mathbf{G}_{t})^{-1}\mathbf{V}_{t}.
\end{equation}
Finally, the attention output $\mathbf{O}_{t}$ is computed as:
\begin{equation}
\mathbf{O}_{t}=\sum_{u}\mathbf{A}_{t,u}\,\mathbf{V}'_{u},\ 
\mathbf{A}_{t,u}=\mathrm{Softmax}\Big(\frac{\mathbf{Q}'_{t}(\mathbf{K}'_{u})^\top}{\sqrt{d}}\Big),
\end{equation}
where $u$ indexes all attended tokens, $d$ denotes the embedding dimension.
By adopting the camera-aware attention, our model strengthens cross-view geometric consistency and enhances semantic reasoning ability in sparse-view settings.

\subsection{Multi-View Depth Estimation}
\label{sec:depth}
\subsubsection{Cost Volume Construction}
Building on camera-aware multi-view color features \(\{\mathbf{F}^c_i\}_{i=1}^N\) extracted in the previous stage, we construct the cost volume using a plane-sweep stereo strategy following MVSplat~\cite{mvsplat}.
For each reference view \(i\), we uniformly sample \(L\) depth candidates \(\{d_m\}_{m=1}^L\) within a predefined near-to-far range. The color feature \(\mathbf{F}^c_j\) from a source view \(j\) is then warped to the reference view \(i\) at each depth candidate \(d_m\), producing a set of warped features \(\{\mathbf{F}^{j \rightarrow i}_{d_m}\}_{m=1}^L\).
We then compute the correlation between these warped features and the original reference feature \(\mathbf{F}^c_i\), and aggregate the results across all source views to form the 3D cost volume for reference view \(i\), denoted as \(\mathbf{C}_i\).

\subsubsection{Depth Regression}
Given the constructed cost volumes $\{\mathbf{C}_i\}_{i=1}^N$ and corresponding Transformer features $\{\mathbf{F}^c_i\}_{i=1}^N$, we predict per-pixel depth map $\{\mathbf{D}_i\}_{i=1}^N$ for each input view.
Specifically, the cost volume and Transformer feature of each view are concatenated and fed into a lightweight 2D CNN U-Net to produce per-pixel depth probability distributions.
The final depth map is then computed as the expectation over \(L\) depth candidates. This per-view depth estimation provides the geometric foundation for predicting Gaussian parameters in the subsequent stage.

\begin{table*}[t]
\centering
\caption{Quantitative comparisons on ScanNet dataset (averaged over 10 scenes). The best results are highlighted in bold.}
\setlength{\tabcolsep}{6pt}
\begin{tabular}{c|cccc|cccc|cccc}
\toprule
\multirow{2.5}{*}{\textbf{Methods}} & \multicolumn{4}{c|}{\textbf{2 Input Views} ($N=2$)} & \multicolumn{4}{c|}{\textbf{3 Input Views} ($N=3$)} & \multicolumn{4}{c}{\textbf{4 Input Views} ($N=4$)} \\
\rule{0pt}{8pt}
& mIoU$\uparrow$ & acc.$\uparrow$ & class acc.$\uparrow$ & FPS$\uparrow$ & mIoU$\uparrow$ & acc.$\uparrow$ & class acc.$\uparrow$ & FPS$\uparrow$ & mIoU$\uparrow$ & acc.$\uparrow$ & class acc.$\uparrow$ & FPS$\uparrow$ \\
\midrule
S-Ray~\cite{S-Ray} & 0.538 & 0.772 & 0.619 & 0.52 & 0.563 & 0.778 & 0.632 & 0.41 & 0.604 & 0.802 & 0.673 & 0.34 \\
\rule{0pt}{7pt}
GSNeRF~\cite{GSNeRF} & 0.529 & 0.751 & 0.616 & 0.25 & 0.550 & 0.752 & 0.648 & 0.23 & 0.587 & 0.796 & 0.671 & 0.19 \\
\rule{0pt}{7pt}
SemGS (Ours) & \textbf{0.754} & \textbf{0.912} & \textbf{0.803} & \textbf{8.49} & \textbf{0.757} & \textbf{0.908} & \textbf{0.811} & \textbf{7.35} & \textbf{0.765} & \textbf{0.919} & \textbf{0.815} & \textbf{6.10} \\
\bottomrule
\end{tabular}
\label{tab:scannet}
\end{table*}

\begin{table*}[t]
\centering
\caption{Quantitative comparisons on ScanNet++ dataset (averaged over 10 scenes). The best results are highlighted in bold.}
\setlength{\tabcolsep}{6pt}
\begin{tabular}{c|cccc|cccc|cccc}
\toprule
\multirow{2.5}{*}{\textbf{Methods}} & \multicolumn{4}{c|}{\textbf{2 Input Views} ($N=2$)} & \multicolumn{4}{c|}{\textbf{3 Input Views} ($N=3$)} & \multicolumn{4}{c}{\textbf{4 Input Views} ($N=4$)} \\
\rule{0pt}{8pt}
& mIoU$\uparrow$ & acc.$\uparrow$ & class acc.$\uparrow$ & FPS$\uparrow$ & mIoU$\uparrow$ & acc.$\uparrow$ & class acc.$\uparrow$ & FPS$\uparrow$ & mIoU$\uparrow$ & acc.$\uparrow$ & class acc.$\uparrow$ & FPS$\uparrow$ \\
\midrule
S-Ray~\cite{S-Ray} & 0.411 & 0.775 & 0.502 & 0.60 & 0.441 & 0.808 & 0.523 & 0.46 & 0.467 & 0.804 & 0.558 & 0.38 \\
\rule{0pt}{7pt}
GSNeRF~\cite{GSNeRF} & 0.421 & 0.760 & 0.503 & 0.29 & 0.450 & 0.794 & 0.538 & 0.26 & 0.479 & 0.813 & 0.562 & 0.21 \\
\rule{0pt}{7pt}
SemGS (Ours) & \textbf{0.625} & \textbf{0.903} & \textbf{0.689} & \textbf{9.24} & \textbf{0.671} & \textbf{0.924} & \textbf{0.734} & \textbf{7.68} & \textbf{0.676} & \textbf{0.927} & \textbf{0.739} & \textbf{6.29} \\
\bottomrule
\end{tabular}
\label{tab:scannet++}
\end{table*}

\subsection{Semantic Reasoning and Gaussian Parameter Prediction}
\label{sec:gaussian-prediction}
Leveraging the extracted multi-view features $\{\mathbf{F}^c_i, \mathbf{F}^s_i\}_{i=1}^N$ and estimated multi-view depth maps $\{\mathbf{D}_i\}_{i=1}^N$, we decode a set of Gaussians to jointly represent the geometry, appearance, and semantics of the scene.
We adopt a dual-Gaussian representation where each pixel of input views corresponds to two Gaussians: a \textit{color Gaussian} for radiance modeling and a \textit{semantic Gaussian} for semantic reasoning. Both Gaussians share geometric attributes while maintaining branch-specific attributes for their respective purposes.

\subsubsection{Shared Geometric Attributes}
Given the per-view depth maps $\{\mathbf{D}_i \in \mathbb{R}^{H \times W} \}_{i=1}^N$ estimated in the previous stage, the 3D Gaussian positions $\{\bm{\mu}_j\}_{j=1}^{N \times H \times W}$ are computed by back-projecting all the pixels to 3D points using the corresponding camera parameters.
The depth probability distribution shares a similar physical meaning to the opacity (points with higher probabilities are more likely on the surface). Accordingly, the Gaussian opacities $\{\alpha_j \in [0,1]\}_{j=1}^{N \times H \times W}$ are predicted using a CNN from the depth probability distributions.

\subsubsection{Branch-Specific Attributes}
The Gaussian positions and opacities implicitly define the 3D scene geometry. Building on these geometric attributes, we reason about other Gaussian attributes from the multi-view Transformer features to model the appearance and semantic.
Given the input images $\{\mathbf{I}_i\}_{i=1}^N$ and color Transformer features $\{\mathbf{F}^c_i\}_{i=1}^N$, a lightweight CNN is used to regress the color coefficients $\{\mathbf{c}_j\}_{j=1}^{N \times H \times W}$ and covariance $\{\mathbf{\Sigma}^c_j\}_{j=1}^{N \times H \times W}$ of \textit{color Gaussians}.
Symmetrically, another lightweight CNN is used to regress the semantic class distribution $\{\mathbf{s}_j\}_{j=1}^{N \times H \times W}$ and covariance $\{\mathbf{\Sigma}^s_j\}_{j=1}^{N \times H \times W}$ of \textit{semantic Gaussians}, conditioned on the input images $\{\mathbf{I}_i\}_{i=1}^N$ and semantic Transformer features $\{\mathbf{F}^s_i\}_{i=1}^N$.

With these estimated attributes, the Gaussians can then be rasterized~\cite{3dgs} to synthesize novel views and semantic maps.
This dual-Gaussian representation jointly encodes the geometry, appearance, and semantics, enabling unified modeling and holistic reasoning of the 3D scene.

\subsection{Training}
In our \textbf{SemGS}, the color branch of the feature extractor and the depth regression 2D CNN are initialized with pre-trained weights from the feed-forward 3DGS model MVSplat~\cite{mvsplat}, which provides reliable priors on scene geometry and appearance. The semantic branch is trained from scratch. Nevertheless, as \textit{semantic Gaussians} and \textit{color Gaussians} share geometric attributes, semantic inference also leverages the geometric priors of the pre-trained model.

To supervise predicted Gaussians, we employ a semantic cross-entropy loss: $\mathcal{L}_{sem} = - \sum_{l}\mathbf{S}^{l} log\hat{\mathbf{S}}^{l}$, where $l$ indexes the semantic classes, $\mathbf{S}^{l}$ is the ground-truth class label, and $\hat{\mathbf{S}}^{l}$ is the predicted class probability distribution, and a color MSE loss: $\mathcal{L}_{c} = \| \mathbf{I} - \hat{\mathbf{I}} \|_2^2$, where $\mathbf{I}$ and $\hat{\mathbf{I}}$ denote the ground-truth and predicted RGB images, respectively.

Semantic models trained solely with the cross-entropy loss often yield predictions that exhibit poor spatial coherence, leading to noisy or irregular outputs within homogeneous regions. Therefore, we design a regional smoothness loss $\mathcal{L}_{rs}$ that enforces the consistency of predicted semantic class distributions between neighboring pixels. Formally, $\mathcal{L}_{rs}$ is computed as:
\begin{align}
\mathcal{L}_{rs} &=
\sum_{l}\sum_{(i,j)\in \mathcal{R}_l} 
\mathbf{1}[(i+1,j)\in\mathcal{R}_l] \|\hat{\mathbf{S}}_{i+1,j}^{\,l}-\hat{\mathbf{S}}_{i,j}^{\,l}\|_1 \nonumber \\
& + 
\sum_{l}\sum_{(i,j)\in \mathcal{R}_l} 
\mathbf{1}[(i,j+1)\in\mathcal{R}_l] \|\hat{\mathbf{S}}_{i,j+1}^{\,l}-\hat{\mathbf{S}}_{i,j}^{\,l}\|_1,
\end{align}
where $\mathcal{R}_l$ denotes the pixel regions belonging to the ground-truth semantic class $l$, and $\hat{\mathbf{S}}_{i,j}^{l}$ denotes the predicted probability distribution at pixel $(i,j)$ for class $l$.
This regional smoothness loss promotes semantic local consistency while preserving sharp inter-class boundaries.

In sum, the final training objective for \textbf{SemGS} is:
\begin{equation}
\mathcal{L} = \lambda_{sem} \mathcal{L}_{sem} + \lambda_c \mathcal{L}_{c} + \lambda_{rs} \mathcal{L}_{rs},
\end{equation}
where $\lambda_{sem}$, $\lambda_c$, and $\lambda_{rs}$ are balancing weights.

\section{EXPERIMENTS}

\subsection{Experimental Setup}
\subsubsection{Implementation Details}
Our \textbf{SemGS} is implemented in PyTorch with a CUDA-based Gaussian rasterizer.
In dual-branch feature extractor, both color and semantic branches share a CNN backbone consisting of 6 CNN-residual blocks, while the semantic branch includes 2 extra residual blocks for semantic-specific refinement. The color branch adopts a Swin Transformer with 6 Transformer blocks (each containing self-attention and cross-attention), whereas the semantic branch adopts 3 Transformer blocks, with feature dimension $d$ set to 128.
The cost volume is refined using a 2D U-Net and regressed into depth maps through a 2-layer CNN. We sample $L=128$ depth candidates.
Branch-specific attributes of color and semantic Gaussians are predicted by separate 2-layer CNNs.
We adopt the Adam optimizer with learning rate of $2\times10^{-4}$.
The loss weights are set as $\lambda_{sem}=0.1$, $\lambda_c=1.0$, and $\lambda_{rs}=0.001$.
Training is conducted on 4 A100 GPUs with batch size 2 per GPU for 300k iterations.

\subsubsection{Baselines \& Datasets}
Notably, only a few prior works have explored generalizable semantic field reconstruction and semantic novel view synthesis from sparse input images. We compare our \textbf{SemGS} with two state-of-the-art methods in this domain: S-Ray~\cite{S-Ray} and GSNeRF~\cite{GSNeRF}. All methods are tested under sparse input settings with $N=2,3,4$ views.

For quantitative evaluation, we adopt ScanNet~\cite{scannet} and ScanNet++~\cite{scannet++} datasets, both of which provide multi-view images, camera poses, and ground-truth semantic annotations. On ScanNet, all methods are trained with 1,000 scenes and tested on 10 held-out scenes, while on ScanNet++ we use 896 training scenes and 10 testing scenes. To further assess the generalization ability, we directly evaluate models trained on ScanNet on unseen domains without finetuning. Specifically, we use Replica~\cite{replica} which contains synthetic scenes and real-world sequences collected by a mobile robot, and conduct qualitative comparisons across all models.

\begin{figure*}
    \centering
    \includegraphics[width=0.98\linewidth]{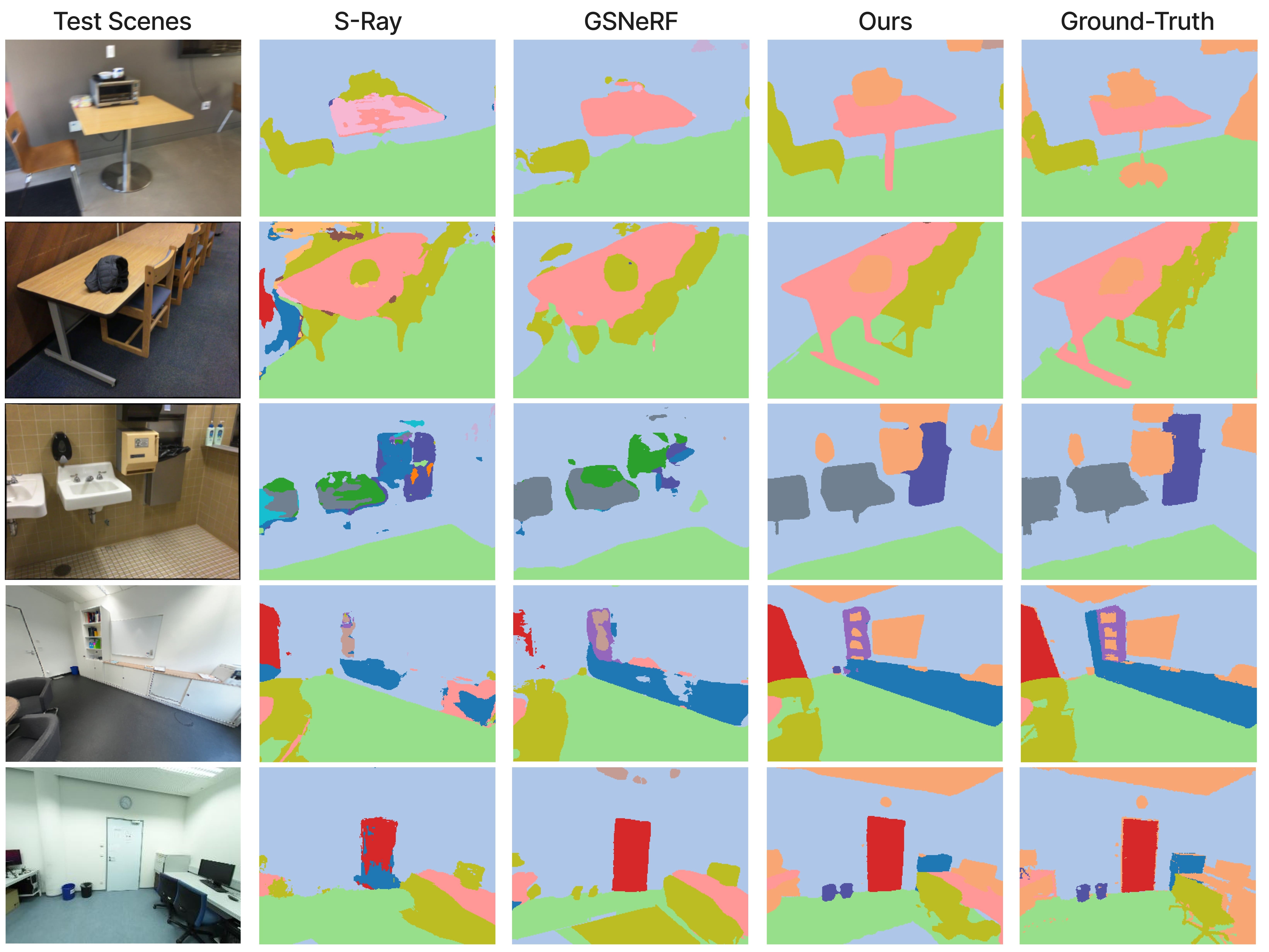}
    \caption{Qualitative comparisons on ScanNet and ScanNet++ datasets. We demonstrate novel-view semantic rendering results of different methods. Compared to existing works, our method produces sharper object boundaries, fewer misclassified regions, and more spatially consistent segmentations.}
    \label{fig:qualitative}
\end{figure*}

\subsubsection{Evaluation Metrics}
Following previous works~\cite{S-Ray, GSNeRF}, we report three semantic metrics: mean Intersection-over-Union (\textit{mIoU}), total pixel accuracy (\textit{acc.}), and average per-class pixel accuracy (\textit{class acc.}). These metrics are computed over the 20 classes defined in the ScanNet benchmark, and comprehensively assess both the overall and class-wise segmentation performance. Additionally, we adopt rendering frames per second (\textit{FPS}) to evaluate the inference speed and computational efficiency of different methods.

\subsection{Quantitative Evaluation}
We quantitatively evaluate our method against S-Ray~\cite{S-Ray} and GSNeRF~\cite{GSNeRF} on ScanNet and ScanNet++ datasets under different numbers of input views. The comparison results are summarized in Tables~\ref{tab:scannet} and~\ref{tab:scannet++}.

On both datasets, \textbf{SemGS} consistently outperforms prior works across all metrics and demonstrates stronger semantic reasoning ability.
With only 2 input views, \textbf{SemGS} already achieves a significant improvement in \textit{mIoU}, showing the effectiveness of our framework under extremely sparse inputs. As the number of views increases, the advantage of \textbf{SemGS} remains pronounced. In terms of pixel accuracy (\textit{acc.}) and class accuracy (\textit{class acc.}), our method also exhibits significant gains, indicating that it produces more reliable semantic maps with enhanced inter-class separability. Overall, these improvements show that our \textbf{SemGS} enables more accurate and robust semantic field reconstruction.

In addition to the semantic accuracy, \textbf{SemGS} also exhibits significantly superior inference efficiency. By leveraging the feed-forward architecture and efficient Gaussian rasterization, our method achieves more than an order of magnitude speedup (see \textit{FPS} metric in Table~\ref{tab:scannet} and Table~\ref{tab:scannet++}) over existing works. This makes \textbf{SemGS} more practical for real-time robotic applications.

\subsection{Qualitative Evaluation}
We present more qualitative comparisons in Fig.~\ref{fig:qualitative}. Overall, our \textbf{SemGS} outperforms existing methods and generates semantic maps that are closer to the ground-truth annotations. 
Specifically, S-Ray often suffers from segmentation errors and blurred boundaries. For example, in the second row of Fig.~\ref{fig:qualitative}, edges of the table are poorly delineated and parts of areas are misclassified as chairs. GSNeRF yields relatively better predictions but tends to produce noisy or fragmented regions, particularly in challenging cases with fine-grained structures, such as the sinks in the third row.

In contrast, our method can produce sharper object boundaries, fewer misclassified regions, and more spatially coherent segmentations. For instance, the table and chairs in the second row are properly separated, and sinks in the third row are distinguished without artifacts. Moreover, in cluttered indoor environments (fourth and fifth rows), \textbf{SemGS} maintains semantic consistency across large planar regions (\textit{e.g.,} walls and floors), while still capturing small objects (\textit{e.g.,} cabinets and trash cans) with high fidelity.

Qualitative results show that our proposed model ensures global semantic consistency while preserving local details. Consequently, \textbf{SemGS} produces more accurate and visually appealing semantic novel views than previous works.

\subsection{Ablation Study}
To validate the effectiveness of each proposed component in \textbf{SemGS}, we conduct ablation studies on ScanNet dataset and the results are reported in Table~\ref{tab:ablation}.

\begin{figure}[t]
    \centering
    \includegraphics[width=0.98\linewidth]{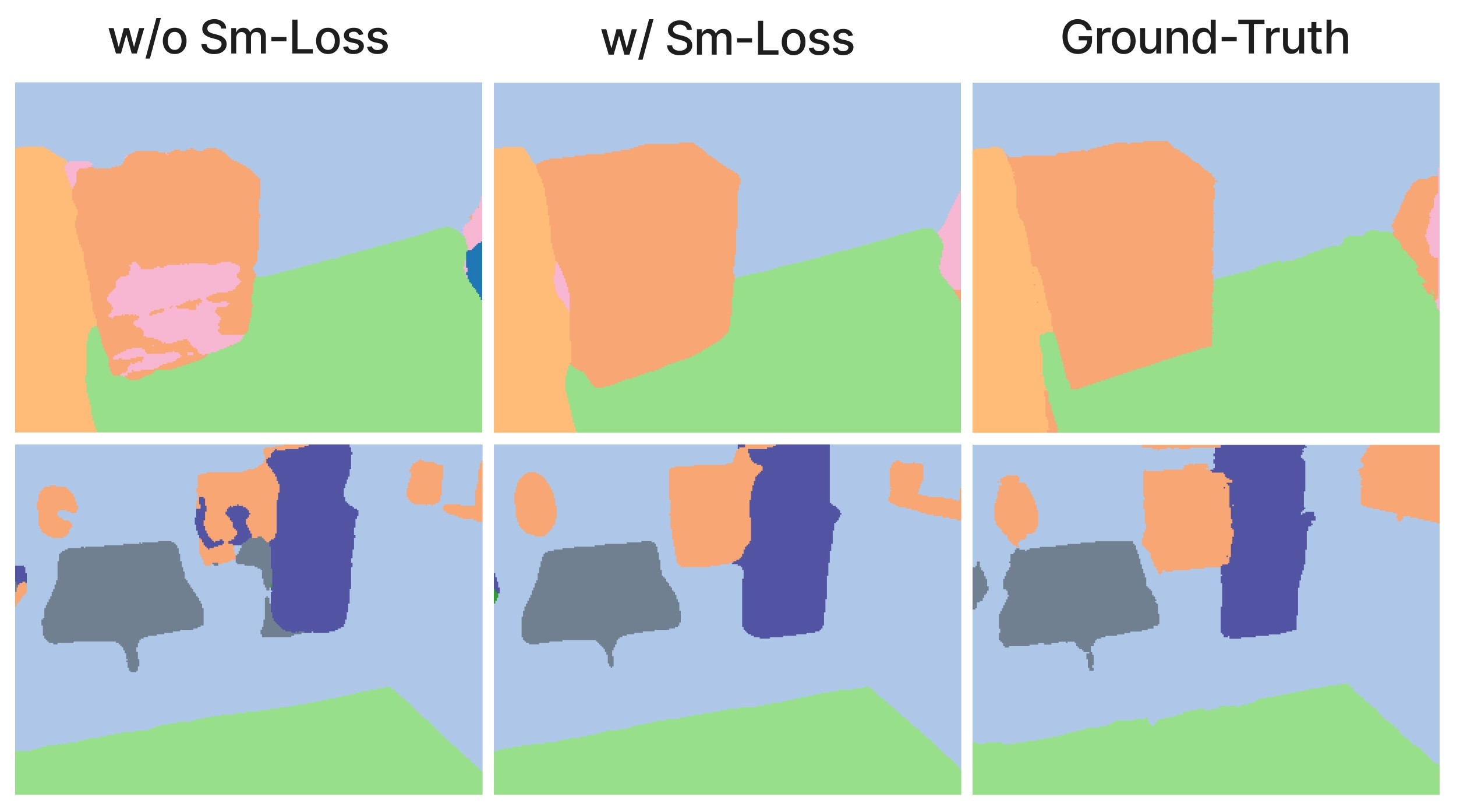}
    \caption{Visualization of the effect of regional smoothness loss (\textit{Sm-Loss}). The proposed loss suppresses noise and enhances semantic coherence.}
    \label{fig:ablation}
\end{figure}

Adopting shared CNN layers (\textit{Sh-Layer}) in feature extractor enhances information exchange between dual branches, yielding noticeable gains in \textit{mIoU} and semantic accuracy.
Replacing vanilla Transformer Blocks with Swin Transformer~\cite{swin} Blocks (\textit{Sw-Trans}) further improves model performance, showing that hierarchical window-based attention is more effective for capturing scene structures.
The proposed camera pose injection mechanism (\textit{C-Inj}) explicitly encodes multi-view camera geometry, which results in better semantic predictions and consistent improvements.
Finally, applying the regional smoothness loss (\textit{Sm-Loss}) regularizes the semantic spatial coherence and reduces local noise. Therefore, it primarily improves pixel accuracy (\textit{acc.}), as this metric is dominated by large-area categories, where the proposed loss can effectively suppress scattered noise and enhance semantic consistency (see Fig.~\ref{fig:ablation}).

Overall, these results validate the individual contributions of each proposed component, and the full model achieves the best performance across all evaluated metrics.

\begin{figure*}
    \centering
    \includegraphics[width=\linewidth]{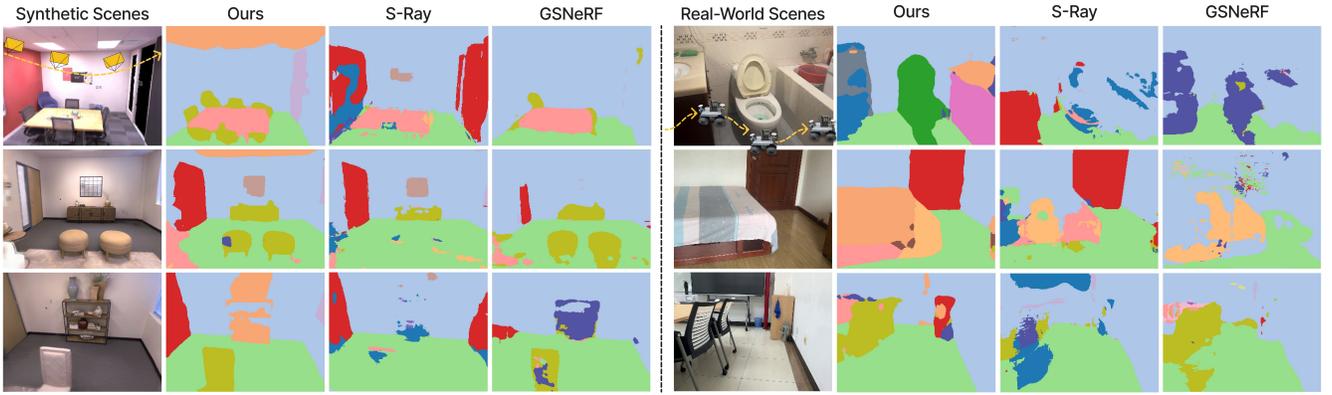}
    \caption{Generalization ability on unseen domains. Different models trained on ScanNet are directly evaluated on Replica synthetic scenes (left) and real-world robot-captured scenes (right). Our \textbf{SemGS} generates more accurate and complete semantic maps than prior works.}
    \label{fig:generalization}
\end{figure*}

\subsection{Generalizability}
To further assess the generalizability of each method, we directly apply models trained on ScanNet to unseen domains, including Replica~\cite{replica} synthetic scenes and real-world robot-captured scenes. The results are shown in Fig.~\ref{fig:generalization}.

In synthetic scenes, our method accurately distinguishes fine-grained structures and preserves sharp boundaries. S-Ray and GSNeRF often produce incomplete or fragmented regions and frequently miss small objects. In real-world robot-captured scenes, S-Ray and GSNeRF suffer from severe noise and misclassify large portions of the scenes, whereas \textbf{SemGS} produces semantic maps that align well with the underlying scene geometry and object layouts.

The superior generalizability of \textbf{SemGS} stems from three key factors. First, the dual-branch feature extractor enables semantic reasoning to leverage low-level structural and textural cues from color features, which remain reliable across domains. Second, the proposed camera-aware attention mechanism explicitly encodes the inter-view camera relations, thereby enhancing the robustness of 3D perception and scene understanding. Third, by sharing geometric attributes between semantic and color Gaussians, we implicitly enforce the geometric consistency. Together, these designs enable our \textbf{SemGS} to achieve superior performance when deployed in previously unseen environments.

\begin{table}[t]
\centering
\caption{Ablation study on ScanNet dataset. The full model, with all components combined, achieves the best performance.}
\renewcommand\arraystretch{1.1}
\setlength{\tabcolsep}{4.5pt}
\begin{tabular}{cccc|ccc}
\toprule
\textit{Sh-Layer} & \textit{Sw-Trans} & \textit{C-Inj} & \textit{Sm-Loss} & mIoU$\uparrow$ & acc.$\uparrow$ & class acc.$\uparrow$ \\
\midrule
\xmark & \xmark & \xmark & \xmark & 0.681 & 0.856 & 0.751 \\
\cmark & \xmark & \xmark & \xmark & 0.718 & 0.873 & 0.776 \\
\cmark & \cmark & \xmark & \xmark & 0.744 & 0.898 & 0.795 \\
\cmark & \cmark & \cmark & \xmark & 0.761 & 0.910 & 0.812 \\
\cmark & \cmark & \cmark & \cmark & \textbf{0.765} & \textbf{0.919} & \textbf{0.815} \\
\bottomrule
\end{tabular}
\label{tab:ablation}
\end{table}

\section{CONCLUSIONS}
In this paper, we present \textbf{SemGS}, a feed-forward framework that constructs generalizable semantic fields from sparse input images.
\textbf{SemGS} employs a dual-branch feature extractor with shared low-level layers, allowing semantic features to leverage the structural and textural cues of color appearance. In addition, we propose to inject camera poses into the Transformer attention mechanism, which enhances the semantic reasoning capability. To further enforce the coherence of semantic predictions, we also design a regional smoothness loss.
Experiments on benchmark datasets show that \textbf{SemGS} not only achieves superior semantic rendering accuracy, but also offers significant improvements in inference speed and generalization ability to unseen environments.
These advantages underscore the potential of our framework for real-world robotic applications.

Despite its strong performance, there still remains room to improve \textbf{SemGS}. First, our framework relies on known cameras (or estimated by off-the-shelf tools~\cite{colmap, vggt}). Inaccuracies in camera poses can propagate into semantic predictions. A future direction is to jointly optimize the camera parameters within the framework, improving robustness to imperfect camera calibration in an end-to-end manner. Second, although \textbf{SemGS} exhibits improved cross-domain generalization, its performance may still degrade when facing drastic domain gaps such as outdoor scenes with highly dynamic objects. Scaling up training with more diverse datasets, combined with strong 2D foundation model features~\cite{clip, sam}, could further enhance robustness in such challenging scenarios. We leave addressing these challenges as promising directions for future work.

\section*{ACKNOWLEDGMENT}
This work was supported by the Natural Science Foundation of China (62332019, 62461160309).


\bibliographystyle{IEEEtran}
\bibliography{IEEEexample}

\end{document}